\pdfoutput=1

\documentclass[11pt]{article}

\usepackage[review]{EMNLP2022}

\usepackage{times}
\usepackage{latexsym}

\usepackage[T1]{fontenc}

\usepackage[utf8]{inputenc}

\usepackage{microtype}
\usepackage{subfigure}
\usepackage{graphicx}
\usepackage{amssymb}
\usepackage{multirow}
\usepackage{pifont}
\usepackage{array}
\usepackage{CJKutf8}
\usepackage{booktabs}
\usepackage{amsmath}
\usepackage{amstext}
\usepackage{enumitem}
\setenumerate[1]{itemsep=0pt,partopsep=0pt,parsep=\parskip,topsep=0pt}
\setitemize[1]{itemsep=0pt,partopsep=0pt,parsep=\parskip,topsep=0pt}
\setdescription{itemsep=0pt,partopsep=0pt,parsep=\parskip,topsep=0pt}

\usepackage{inconsolata}

\title{DialogConv: A Lightweight Fully Convolutional Network for Multi-view Response Selection}

\author{Yongkang Liu\textsuperscript{\rm 1}, Shi Feng\textsuperscript{\rm 1}, Wei Gao\textsuperscript{\rm 2}, Daling Wang\textsuperscript{\rm 1} and Yifei Zhang\textsuperscript{\rm 1} \\
\textsuperscript{\rm 1} Northeastern University, China\\
\textsuperscript{\rm 2} Singapore Management University, Singapore\\
\texttt{misonsky@163.com, fengshi@cse.neu.edu.cn}\\
\texttt{weigao@smu.edu.sg, \{wangdaling,zhangyifei\}@cse.neu.edu.cn }
}

\begin{document}
\maketitle
\begin{abstract}
Current end-to-end retrieval-based dialogue systems are mainly based on Recurrent Neural Networks or Transformers with attention 
mechanisms. Although promising results have been achieved, these models often suffer from slow inference or huge number of parameters. In this paper, we propose a novel lightweight fully convolutional architecture, called DialogConv, for response selection. DialogConv is exclusively built on top of convolution to extract matching features of context and response. Dialogues are modeled in 3D views, where DialogConv performs convolution operations on embedding view, word view and utterance view to capture richer semantic information from multiple contextual views. On the four benchmark datasets, compared with state-of-the-art baselines, DialogConv is on average about $8.5\times$ smaller in size, and $79.39\times$ and $10.64\times$ faster on CPU and GPU devices, respectively. At the same time, DialogConv achieves the competitive effectiveness of response selection.
\end{abstract}

\section{Introduction}
An important challenge in building intelligent dialogue systems is the response selection problem, which aims to select an appropriate response from a set of candidates given a dialogue context. Such retrieval-based dialogue systems have attracted great attention from academia and industry due to the advantages of informative and fluent responses produced~\cite{tao2021survey}.

\begin{figure}[t] 
\centering
\includegraphics[width=0.85\columnwidth]{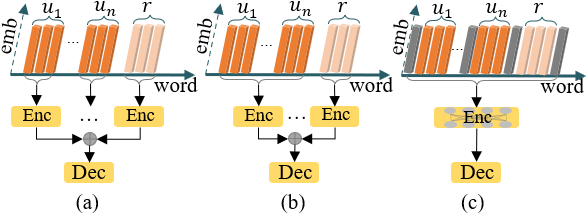}
\caption{Flat modeling. (a) is separate pattern, (b) is concatenated pattern and (c) is PrLM pattern. Grey bars in (c) are embedded representations of special symbols.}
\label{multi-view}
\end{figure}

The existing retrieval-based dialogue systems can be divided into three patterns according to the way of input handling~\cite{zhang2021advances}: (i) \textit{Separate Pattern}~\cite{wu2017sequential,zhang2018modeling,Zhou2018,gu2019interactive}; (ii) \textit{Concatenated Pattern}~\cite{tan2015lstm,zhou2016multi}; (iii) \textit{PrLM (Pretrained Language Model) Pattern}~\cite{cui2020mutual,gu2020speaker,liu2021graph}. Separate Pattern (i.e., Figure~\ref{multi-view} (a)) encodes utterances individually, while Concatenated Pattern (i.e., Figure~\ref{multi-view} (b)) concatenates all utterances into a continuous word sequence. Methods based on these two patterns usually have Recurrent Neural Networks (RNNs)~\cite{hochreiter1997long,cho2014learning} and attention mechanism~\cite{bahdanau2014neural} as the backbone. Although promising results have been achieved, these methods are generally slow in training and inference due to their recurrent nature.

\begin{figure*}[t]
\centering
\includegraphics[width=0.85\textwidth]{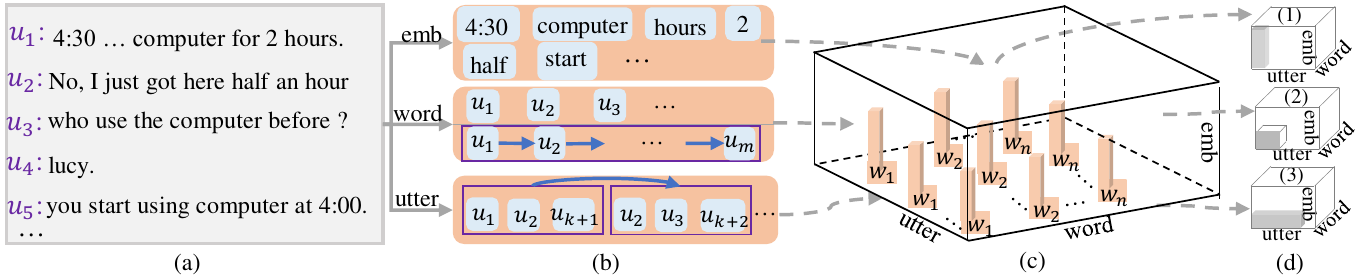}
\caption{Stereo view modeling. (a) An example of multi-turn dialogue; (b) Features from different views; (c) A schematic diagram of stereo view; (d) Convolution on different views ((1) is convolution in embedding view; (2) is convolution in word view; and (3) is convolution in utterance view)}
\label{example}
\end{figure*}

The PrLM Pattern (i.e., Figure~\ref{multi-view} (c)) uses special symbols to connect all utterances into a continuous sequence, similar to Concatenated Pattern. While PrLM Pattern has obtained state-of-the-art performance in response selection~\cite{cui2020mutual,gu2020speaker,liu2021graph}, this method having Transformer~\cite{vaswani2017attention} as the de facto standard architecture suffer from a large number of parameters and heavy computational cost. Very large models not only lead to increased training costs, but also prevent researchers from iterating quickly. At the same time, slow inference hinders development and deployment of dialogue systems in real-world scenarios.

Furthermore, these three patterns treat dialogue contexts as flat structures~\cite{li2021conversations}. Methods based on such flat structures usually capture the sequential features of text by considering each word as a unit. 
However, previous work~\cite{lu2019constructing} revealed that given a multi-turn dialogue (e.g., Figure~\ref{example} (a)), the context of the dialogue can exhibit a composition of 3D stereo structures as we view utterances in each dimension (shown as Figure~\ref{example} (b) and (c))). As shown in Figure~\ref{example} (b), the embedding view can represent features of each individual word, the word view can represent the features from the whole conversation and a single utterance, and the utterance view can capture the dependencies between different localities composed of adjacent utterances. Existing methods~\cite{gu2019interactive,zhou2016multi,gu2020speaker} only extract features based on the flat structures, but cannot simultaneously capture complex features from such stereoscopic views.

In this paper, we propose a \textit{lightweight} fully\footnote{Here ‘fully’ means DialogConv is built exclusively on CNNs.} convolutional network model, called DialogConv, without any RNN and attention module for multi-view response selection. Different from previous studies~\cite{zhou2016multi,gu2019interactive,gu2020speaker,li2021conversations} which model the dialogue in a flat view, DialogConv models the dialogue context and response together in the 3D space of the stereo views, i.e., embedding view, word view, and utterance view (as shown in Figure~\ref{example} (d)). In the embedding view, the word-level features will be refined through convolution operations on the plane formed by the word sequence dimension and the utterance dimension. In the word view, the global conversation features  will be captured by concatenating all words into a continuous sequence, and the features of each utterance will be refined by performing convolution on each utterance. In the utterance view, the dependency features between different local contexts will be distilled by performing convolution across different utterances. In general, DialogConv can simultaneously extract features with different granularities from the stereo structure.

DialogConv is completely based on CNN, which uses much fewer parameters and computing resources. DialogueConv has an average number of parameters of 12.4 million, which is on average about $8.5 \times$ smaller than other models. The inference speed of DialogConv can be 
on average about $79.39 \times$ faster on CPU device and $10.64 \times$ faster on GPU device than existing models. Moreover, DialogConv achieves competitive results on four benchmarks and performs even better when pre-trained with contrastive learning. In summary, we make the following contributions:
\begin{itemize}[leftmargin=*]
    \item We propose an efficient convolutional response selection model, DialogConv, which, to our best knowledge, is the first response selection model built entirely on multiple convolutional layers without any RNN or attention module. 
    \item We model dialogue from stereo views, where 2D and 1D convolution operations are performed on embedding, word and utterance views, and thus DialogConv can capture features from stereo views simultaneously. 
    \item Extensive experiments on four benchmark datasets show that DialogConv with fewer parameters can achieve competitive performance with faster speed and less computing resources. The code is available o Github\footnote{https://github.com/misonsky/DialogConv}.
\end{itemize}

\section{Related Work}
\subsection{Retrieval-based Dialogue System}
Most existing retrieval-based dialogue systems~\cite{wu2017sequential,gu2019interactive,liu2021graph} focus on \textit{matching} between dialogue context and response. These methods attempt to mine deep semantic features through sequence modeling, e.g., using attention-based pairwise matching mechanisms to capture interaction features between dialogue context and candidate response. However, previous research~\cite{sankar2019neural,li2021conversations} shows that these methods fail to fully exploit the conversation history information. In addition, methods based on recurrent neural network suffer from slow inference speed due to the nature of recurrent structures. Although Transformer-based methods~\cite{vaswani2017attention} get rid of the weakness of recurrent structure, they are usually plagued by a large number of parameters~\cite{wu2019lite}, making the training and inference of Transformer-based models require a lot of computational cost.  In this paper, we propose a multi-view approach to model dialogue context based on a fully convolutional structure and a lightweight model that is smaller and faster than most existing methods.

\subsection{Convolutional Neural Networks (CNN)}
For the past few years, CNNs have been the go-to model in computer vision. The main reason is that CNN enjoys the advantage of parameter sharing and is better at modeling local structures. A large number of excellent architectures based on CNN have been proposed~\cite{krizhevsky2012imagenet,he2016deep,dai2021coatnet}. For text processing, convolutional structures are good at capturing local dependencies of text and are faster than RNNs~\cite{hochreiter1997long}. Therefore, some studies~\cite{wu2016sequential,lu2019constructing,yuan2019multi} employ convolutional structures to aggregate the matching features between dialogue contexts and responses. However, these works usually require combining attention mechanisms or the skeleton structure of RNN with CNNs. Furthermore, these studies treat dialogue context as a flat structure. In this paper, we propose a novel fully convolutional architecture to extract matching features from stereo views, which can simultaneously extract the features with different granularities from different views.

\begin{figure*}[t]
\centering 
\includegraphics[width=0.85\textwidth]{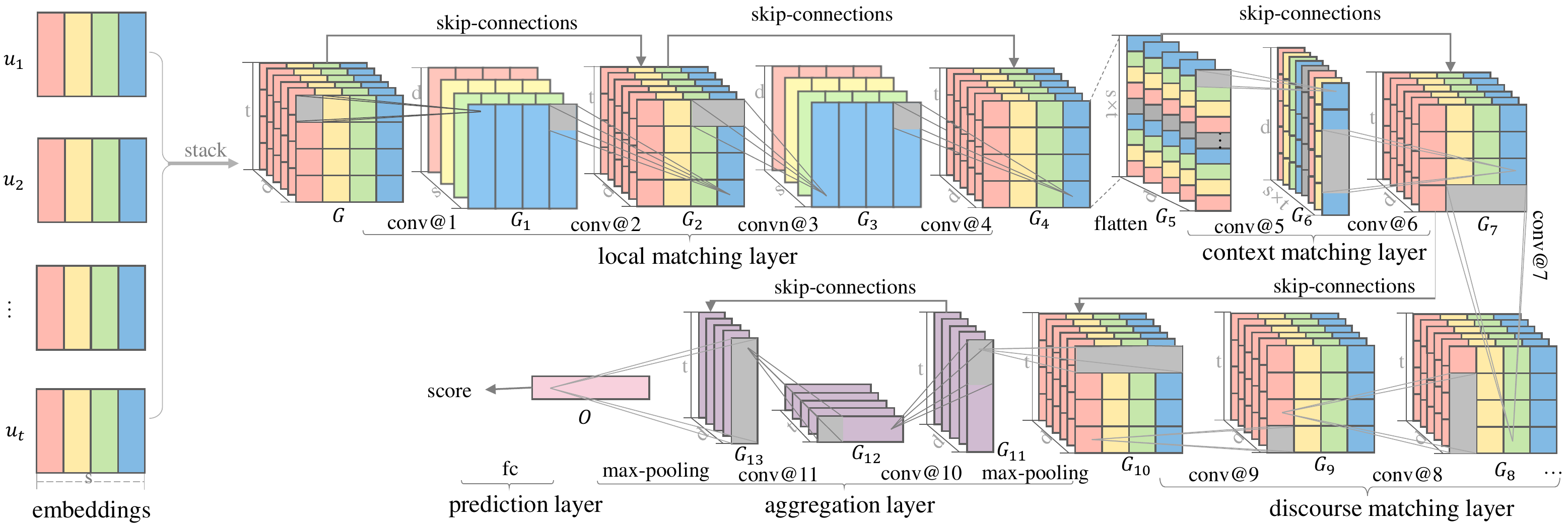}
\caption{Overview of our DialogConv. The conv@\textit{i} symbol represents the $i$-th convolution operation.}
\label{framework}
\end{figure*}

\section{Methodology}
\subsection{Problem Formulation}
In this paper, an instance in the dialogue dataset can be represented as $(\boldsymbol{C},y)$, where $\boldsymbol{C}$=$(\boldsymbol{u_1},\boldsymbol{u_2},\dots,\boldsymbol{u_{t-1}},\boldsymbol{r})$ represents the set of dialogue contexts $(\boldsymbol{u_1},\boldsymbol{u_2},\dots,\boldsymbol{u_{t-1}})$ and the response $\boldsymbol{r}$, $\boldsymbol{u_i}$ is the $i$-th utterance, and $y \in \{0,1\}$ is the class label of $\boldsymbol{C}$. As the core of retrieval-based dialogue system, the purpose of response selection is to build a discriminator $g(\boldsymbol{C})$ on $(\boldsymbol{C},y)$ to measure the matching between the dialogue context and response.

\subsection{Fully Convolutional Matching}
We propose a fully convolutional encoder for multi-view response selection. Multiple views include embedding view, word view, and utterance view. In the embedding view, the convolution operations are performed on the plane formed by the word sequence dimension and the utterance dimension, and word-level features can be extracted through nonlinear transformations between different embeddings. In the word view, global dialogue context features will be captured by convolution of a contiguous sequence connecting all words, and features of each utterance will be obtained by performing convolution on each utterance. In the utterance view, DialogConv is responsible for capturing the dependency features between different local contexts composed of adjacent utterances. Figure~\ref{framework} shows an overview of our proposed DialogConv, which consists of six layers: (i) embedding layer; (ii) local matching layer; (iii) context matching layer; (iv) discourse matching layer; (v) aggregation layer; (vi) prediction Layer.

\emph{Symbol Definition}: The embedding layer uses a pretrained word embedding model to map each word in $\boldsymbol{C}$ to a vector space. We stack  $\boldsymbol{C}$ chronologically into a 3D tensor $\boldsymbol{G} \in \mathbb{R}^{t\times \ell \times d}$, where $d$ represents the dimension of word embedding, $\ell$ represents the length of the utterance and $t$ is the number of utterances including the response. $\boldsymbol{G}$ is the input to DialogConv. We use Conv2D$_{k \times s}^v$ and Conv1D$_w^v$ to denote the convolution operations, where Conv2D$_{k \times s}^v$ denotes a two-dimensional convolution with a convolution kernel size of $k \times s$, Conv1D$_w^v$ represents a one-dimensional convolution with a convolution kernel size of $w$, and $v$ represents a specific view. We will describe the details of the remaining layers in the following subsections.

\subsubsection{Local Matching Layer}
The local matching layer is responsible for extracting features of each utterance. The local matching stage contains features from the embedding and word views. Firstly, we employ $1\times 1$ convolutions in the embedding view and the word view, respectively. The process can be formally described as:
\begin{eqnarray}
\boldsymbol{G}_1 & = & \textrm{Conv2D}_{1 \times 1}^{embedding}(\sigma(\boldsymbol{G})) \\
\boldsymbol{G}_2 & = & \textrm{Conv2D}_{1 \times 1}^{word}(\boldsymbol{G}_1) + \boldsymbol{G} 
\end{eqnarray}
where $\sigma(\cdot)$ stands for GELUs activation function~\cite{hendrycks2016gaussian}. The $1 \times 1$ convolution pays attention to the information of the current element itself and does not consider the influence of local context. The features of individual words will be extracted from the embedding and word views by $1 \times 1$ convolution. Multi-scale convolution~\cite{szegedy2015going,gao2019res2net} has been shown to be effective in extracting local features. Therefore, we use a $k_1$×$s_1$ convolution in the word view and a $1 \times 1$ convolution in the embedding view to capture the matching features of each utterance. The formal description is given as follows:
\begin{eqnarray}
\boldsymbol{G}_3 & = & \textrm{Conv2D}_{k_1 \times s_1}^{word}(\sigma(\boldsymbol{G}_2))
\\
\boldsymbol{G}_4 & = & \textrm{Conv2D}_{1 \times 1}^{embedding}(\boldsymbol{G}_3) + \boldsymbol{G}_2
\end{eqnarray}
Note that we focus on the features of a single utterance in the local matching layer.

\subsubsection{Context Matching Layer}
The context matching layer is responsible for extracting matching features of the global dialogue context.
Firstly, we flatten $\boldsymbol{G}_4$ into a two-dimensional tensor $\boldsymbol{G}_5 \in \mathbb{R}^{(t \times \ell) \times d}$. This is equivalent to concatenating all utterances in chronological order into one continuous sequence of words. Then, we use convolution across words sequence with kernel size of $w_1$ in the embedding view, and kernel size of $w_2$ in the word view. Details are as follows:
\begin{eqnarray}
\boldsymbol{G}_6 & = & \textrm{Conv1D}_{w_1}^{embedding}(\sigma(\boldsymbol{G}_5))
\\
\boldsymbol{G}_7 & = & f_{reshape}(\textrm{Conv1D}_{w_2}^{word}(\boldsymbol{G}_6)) + \boldsymbol{G}_5
\end{eqnarray}
where $f_{reshape}$ is a function that reshapes the output of the convolution to the same shape as $\boldsymbol{G}_5$ and $\boldsymbol{G}_7 \in \mathbb{R}^{t \times \ell \times d}$. The features of the global dialogue context can be aggregated by a nonlinear transformation between different words concatenating all utterances. The features of the global context are basis for extracting the dependency features between different local contexts.

\subsubsection{Discourse Matching Layer}
The discourse matching layer is responsible for capturing the dependencies between different local contexts composed of adjacent utterances. Modeling dependency features is beneficial for capturing changes in implicit topics, intentions, etc. in the dialogue context, which is important for choosing the correct response. We employ orthogonal convolution to extract dynamic dialogue flow features across utterances to capture discourse-level changes. The specific process is formulated as follows:
\begin{eqnarray}
\boldsymbol{G}_8 & = & \textrm{Conv2D}_{1 \times s_2}^{utterance}(\sigma(\boldsymbol{G}_7))
\\
\boldsymbol{G}_9 & = & \textrm{Conv2D}_{s_2 \times 1}^{utterance}(\boldsymbol{G}_8)
\\
\boldsymbol{G}_{10} & = & \textrm{Conv2D}_{1 \times 1}^{utterance}(\boldsymbol{G}_9) + \boldsymbol{G}_7
\end{eqnarray}
where the $1 \times s_2$ convolution and $s_2 \times 1$ convolution are called orthogonal convolutions because the direction of their convolution kernels is vertical. The $1 \times s_2$ convolution is responsible for forming semantic flow based on the context-level features of a single utterance, and the $s_2 \times 1$ convolution extracts discourse structural features according to the depth of dialogue. Finally, we integrate features of utterances through the $1 \times 1$ convolution.

\subsubsection{Aggregation Layer}
The aggregation layer is responsible for obtaining high-level semantic information by integrating the matching features from previous layers. First, we use max-pooling to obtain the sentence representation $\boldsymbol{G}_{11} \in \mathbb{R}^{t \times d}$. Then, we employ two layers of convolution to extract matching features along the embedding dimension in the embedding view and the depth of dialogue in the utterance view, respectively. The formulation is as follows:
\begin{eqnarray}
\boldsymbol{G}_{12} & = & \textrm{Conv1D}_{w_3}^{embedding}(\boldsymbol{G}_{11})
\\
\boldsymbol{G}_{13} & = & \textrm{Conv1D}_{w_4}^{utterance}(\boldsymbol{G}_{12}) + \boldsymbol{G}_{11}
\end{eqnarray}
where $w_3$ and $w_4$ are the convolution kernel sizes.
We again use a max-pooling operation based on $\boldsymbol{G}_{13}$ to obtain the final context representation $\boldsymbol{O}$.

\subsection{Self-supervised Pre-training}
As a lightweight neural structure, the performance of DialogConv can be further improved by a pre-training strategy using a small corpus. While the masked language model pretraining~\cite{devlin2018bert,Lan2020} usually requires large-scale corpora, self-supervised contrastive learning can generally learn representations with a relatively small-scale corpus. Therefore, we employ contrastive learning to learn effective representations by pulling semantically close neighbors together and pushing apart non-neighbors~\cite{1640964}. Given a set of paired examples $D={(x_i,x_i^+)}$, where $x_i$ is the dialogue context $c$ and $x_i^+$ is the correct response. We adopt the previous contrastive learning framework~\cite{liu2021simcls} and employ a cross-entropy objective, where the negatives $x_i^-$ include responses with $y=0$ and in-batch negatives~\cite{chen2017sampling}. The training objective is:
\begin{equation}
L = \log \frac{e^{sim(x_i,x_i^+) / \tau}}{\sum_{j=1}^{|x_i^-|}e^{sim(x_i,x_{ij}^-) / \tau} + e^{sim(x_i,x_i^+) / \tau}}
\end{equation}
where $\tau$ is a temperature hyperparameter, $x_{ij}^-$ represents the $j$-th negative example of $x_i$, $x_i^+$ represents the positive example of $x_i$, and $sim(\cdot,\cdot)$ is the cosine similarity.

\begin{table*}[t]
\centering
\small
\begin{tabular}{l|ccc|cccccc}
\toprule
   & \multicolumn{3}{c}{Ubuntu (English)} & \multicolumn{6}{|c}{Douban (Chinese)} \\ \midrule
   Method & R10@1 & R10@2 & R10@5 & MAP &MRR & P@1 & R10@1 &R10@2 & R10@5  \\ \midrule
   MV-LSTM & 0.653 & 0.804 & 0.946 & 0.498 & 0.538 & 0.348 & 0.202 & 0.351 & 0.710  \\
   MH-LSTM & 0.653 & 0.799 & 0.944 & 0.500 & 0.537 & 0.345 & 0.202 & 0.348 & 0.720  \\ \midrule
   DUA & 0.757 & 0.868 & 0.962 & 0.551 & 0.599 & 0.421 & 0.243 & 0.421 & 0.780  \\ 
   DAM & 0.767 & 0.874 & 0.969 & 0.550 & 0.601 & 0.427 & 0.254 & 0.410 & 0.757 \\ 
   MRFN & 0.786 & 0.886 & 0.976 & 0.571 & 0.617 & 0.448 & 0.276 & 0.435 & 0.783 \\ 
   IMN & 0.794 & 0.889 & 0.974 & 0.570 & 0.615 & 0.433 & 0.262 & 0.452 & 0.789  \\ 
   IoI & 0.796 & 0.894 & 0.974 & 0.573 & 0.621 & 0.444 & 0.269 & 0.451 & 0.786 \\
   MSN & 0.800 & \underline{0.899} & \underline{0.978} & \underline{0.587} & 0.632 & \textbf{0.470} & \textbf{0.295} & 0.452 & 0.788\\ \midrule
   BERT-2-128 & 0.647 & 0.767 & 0.911 & - & - & - & - & - & -  \\ 
   BERT-4-256 & 0.706 & 0.809 & 0.932 & - & - & - & - & - & -  \\ 
   BERT-4-512 & 0.757 & 0.869 & 0.966 & - & - & - & - & - & -  \\ 
   BERT-8-512 & 0.788 & 0.888 & \underline{0.978} & - & - & - & - & - & -  \\
   BERT-12-768 & \textbf{0.808} & 0.897 & 0.975 & \textbf{0.591} & \underline{0.633} & 0.454 & 0.280 & \textbf{0.470} & \textbf{0.828} \\
   DBERT-6-768 &0.783 & 0.879 & 0.968 & 0.542 & 0.592 & 0.418 & 0.249 & 0.407 & 0.765 \\
   TBERT-4-312 &0.638 & 0.766 & 0.922 & 0.532 & 0.567 & 0.378 & 0.235 & 0.397 & 0.742 \\
   TBERT-6-768 &0.729 & 0.835 & 0.954 & 0.559 & 0.597 & 0.413 & 0.257 & 0.417 & 0.796 \\ \midrule
  DialogConv & 0.788 & 0.883 & \textbf{0.979} & 0.571 & 0.624 & 0.432 & 0.272 & \underline{0.453} & 0.785 \\
  DialogConv* & \underline{0.801} & \textbf{0.904} & 0.976 & 0.572 & \textbf{0.634} & \underline{0.457} & \underline{0.282} & 0.452 & \underline{0.825}\\ \bottomrule
\end{tabular}
\caption{Results on Ubuntu and Douban datasets. The first, second and third groups of models belong to the Concatenated Pattern, Separate Pattern and PrLM-based Pattern, respectively. DialogConv* represents the performance when pretraining with contrastive learning. \textbf{Bold} indicates the best result, and \underline{underline} indicates the second best result. $X$ represents the number of layers and $Y$ represents the hidden size of the model in BERT-$X$-$Y$, DBERT-$X$-$Y$ and TBERT-$X$-$Y$. TBERT stands for TinyBERT, and DBERT stands for DistilBERT. The `-' indicates no corresponding BERT version is available.}
\label{compared_results}
\end{table*}

\section{Experiments and Results}
The baselines are described in the Appendix~\ref{baseline}.
\subsection{Datasets}
\label{dataset}
We conduct extensive experiments on four public datasets: (i) Ubuntu Dialogue (Ubuntu)~\cite{lowe2015ubuntu}; (ii) Multi-Turn Dialogue Reasoning (MuTual)~\cite{cui2020mutual}; (iii) Douban Conversation Corpus (Douban)~\cite{wu2016sequential}; (iv) E-commerce Dialogue Corpus (ECD)~\cite{zhang2018modeling}. \textbf{Ubuntu} consists of 1 million context-response pairs for training, 0.5 million pairs for validation, and 0.5 million pairs for testing. The ratio of the positive and the negative is 1:1 for training, and 1:9 for validation and testing. \textbf{Douban} consists of 1 million context-response pairs for training, 50k pairs for validation, and 10k pairs for testing. Response candidates are retrieved from Sina Weibo and labeled by human judges. \textbf{ECD} contains 1 million context-response pairs for training, 10k pairs for validation, and 10k pairs for testing and consists of five different types of conversations (e.g., commodity consultation, logistics express, recommendation, negotiation and chitchat) based on over 20 commodities. \textbf{MuTual} is the first human-labeled reasoning-based dataset for multi-turn dialogue, which contains 7,088 context-response pairs for training, 886 pairs for validation, and 886 pairs for testing. The ratio of the positive and the negative is 1:3 in the training, validation and test sets.
\begin{table}[t]
\centering
\scriptsize
\begin{tabular}{p{1.45cm}|p{0.55cm}p{0.55cm}p{0.55cm}|p{0.5cm}p{0.5cm}p{0.5cm}}
\toprule
   & \multicolumn{3}{c}{ECD (Chinese)} & \multicolumn{3}{|c}{MuTual (English)} \\ \midrule
   Method & R10@1 & R10@2 & R10@5 & R@1 & R@2 & MRR   \\ \midrule
   MV-LSTM & 0.412 & 0.591 & 0.857 & \quad- & \quad- & \quad-  \\
   QANET & 0.455 & 0.662 & 0.920 & 0.247 & 0.517  & 0.522 \\
   BIDAF & 0.491 & 0.708 & 0.933 & 0.357 & 0.589 & 0.589 \\
   MH-LSTM & 0.410 & 0.590 & 0.858 & \quad- & \quad- & \quad- \\ \midrule
   DL2R & 0.399 & 0.571 & 0.842 & \quad- & \quad- & \quad- \\ 
   DUA & 0.501 & 0.700 & 0.921 & 0.437 & 0.698 & 0.658 \\ 
   DAM & 0.526 & 0.727 & 0.933 & 0.458 & 0.718 & 0.673 \\
   IMN & 0.621 & 0.797 & 0.964 & 0.404 & 0.622 & 0.638 \\ 
   IoI & 0.563 & 0.768 & 0.950 & 0.421 & 0.686 & 0.647 \\
   MSN & 0.606 & 0.770 & 0.937 & 0.420 & 0.677 & 0.646 \\ \midrule
   BERT-2-128 & \quad- & \quad- & \quad- & 0.520 & 0.765 & 0.715 \\ 
   BERT-4-256 & \quad- & \quad- & \quad- & 0.558 & 0.800 & 0.742 \\ 
   BERT-4-512 & \quad- & \quad- & \quad- & 0.607 & 0.837 & 0.772 \\ 
   BERT-8-512 & \quad- & \quad- & \quad- & 0.619 & 0.816 & 0.774 \\
   BERT-12-768 & 0.610 & 0.814 & \textbf{0.973} & \underline{0.648} & \underline{0.847} & \textbf{0.795} \\
   DBERT-6-768 & 0.517 & 0.695 & 0.885 & 0.602 & 0.836 & 0.769 \\
   TBERT-4-312 & 0.449 & 0.583 & 0.854 & 0.534 & 0.778 & 0.724 \\
   TBERT-6-768 & 0.587 & 0.794 & 0.953 & 0.615 & 0.833 & 0.785 \\ \midrule
  DialogConv & \underline{0.827} & \underline{0.889} & 0.962 & 0.602 & 0.834 & 0.769  \\
  DialogConv* & \textbf{0.844} & \textbf{0.891} & \underline{0.963} & \underline{0.622} & \underline{0.854} & \underline{0.782} \\ \bottomrule
\end{tabular}
\caption{Results on ECD and MuTual datasets. The `-' indicates no corresponding BERT version is available.}
\label{compared_results_2}
\end{table}
\begin{table*}[t]
\centering
\small
\begin{tabular}{r|p{1.6cm}p{1.6cm}p{1.6cm}p{1.6cm}|p{0.8cm}p{0.8cm}p{0.8cm}p{0.9cm}}
\toprule
   \multirow{3}{*}{Model} &
   \multicolumn{4}{c}{Inference Time (CPU/GPU)} & 
   \multicolumn{4}{|c}{Parameters (M)} \\
   \cline{2-5} \cline{5-9} & Ubuntu (m) & Douban (s) & ECD (s) & MuTual (s) & Ubuntu & Douban & ECD & MuTual \\
   \midrule
   DAM & 177/45 & 227/68 & 227/66 & 91/38 &95 &67 &13 &8 \\
   DUA & 143/49 & 176/64 & 175/64 & 64/26 &96 &70 &16 &15 \\
   IOI & 347/39 & 421/49 & 429/47 & 157/22 &96 &69 &15 &10\\
   MSN & 105/13 & 128/17 & 126/14 & 45/7 &89 &62 &11 &13 \\ \midrule
   BERT-2-128 & 54/14 & \quad- & \quad- & 24/9 &4 &\quad- &\quad- &4 \\ 
   BERT-4-256 & 352/37 & \quad- & \quad- & 81/12 &11 &\quad- &\quad- &11 \\
   BERT-4-512 & 1372/53 & \quad- & \quad- & 339/24 &29 &\quad- &\quad- &29 \\
   BERT-8-512 & 3174/82 & \quad- & \quad- & 667/42 &41 &\quad- &\quad- &41 \\
   BERT-12-768 & 8991/219 & 4922/240 & 4681/239 & 1694/98 &110 &102 &102 &110 \\
   DBERT-6-768 & 1339/83 & 1591/92 & 1612/87 & 595/33 & 67 & 54 & 54 & 67 \\
   TBERT-4-312 & 376/34 & 450/35 & 445/30 & 165/13 & 14 & 11 & 11 & 14 \\
    \hline
   DialogConv & \textbf{13/5} & \textbf{18/7} & \textbf{16/7} & \textbf{7/3} &\textbf{23} &\textbf{13} &\textbf{9} &\textbf{4} \\ \bottomrule
\end{tabular}
\caption{Comparison of model inference time and the scale of parameters. "m" ("s") stands for minutes (seconds). The number of parameters of Chinese and English BERT is different because their vocabularies differ. The `-' indicates no corresponding BERT version is available.}
\label{ppl}
\end{table*}

\subsection{Evaluation Metrics}
We follow previous research~\cite{zhang2021advances} using evaluation metric Rn@k to measure model performance on the datasets Ubuntu, Douban and ECD, which calculates the proportion of truly positive responses among the top-$k$ responses selected from a list of $n$ available candidates for a context. In addition, the traditional metrics MAP (Mean Average Precision)~\citep{Ricardo2016} and MRR (Mean Reciprocal Rank)~\citep{voorhees1999trec} are employed on Douban. We use recall at position 1 of 4 candidates (R@1), recall at position 2 of 4 candidates (R@2) and MRR on MuTual dataset, following previous study~\cite{liu2021graph}. The Ubuntu, Douban and ECD test sets provide ten candidate responses, while the MuTual provides four candidate responses, leading them to adopt different evaluation metrics.

\subsection{Implementation Details}
\label{ImpDe}
\textbf{Model Details}: We implement DialogConv using Tensorflow 2 and train DialogConv on a server with an Intel(R) Core(TM) i7-10700 CPU 2.90HZ and a single GeForce RTX 2070 SUPER GPU (8G).  In experiments, we consider up to 10 turns and 50 words for the Ubuntu, Douban and ECD datasets, and up to 8 turns and 50 words for the MuTual dataset. The dimension of word embeddings is set to 200. We set the convolution filter sizes $k_1=1, w_1=1, w_2=5, w_3=3, w_4=1, s_1=3 \text{ and } s_2=3$. Layers 1, 2, 3, 4, 8, 9 and 10 use 2D convolutions, and layers 6, 7, 12 and 13 use 1D convolutions. We set the stride of all convolutional layers to $[1,1]$ or 1. The filter size of convolution layers 1, 2, 4, 5, 9 and 11 is set to $[1,1]$. The filter sizes of the convolution layers 3, 6, 7, 8 and 10 are set to $[1,3], 5, [1,3], [3,1] \text{ and } 3$, respectively. 

\textbf{Self-supervised Pre-training}: We conduct  small-scale pretraining on the training set of downstream tasks through contrastive learning, such as Ubuntu and Douban. Negative instances include not only negative examples provided by the dataset, but also candidate responses from other instances in the same batch. We use the Stochastic Gradient Descent (SGD) optimizer~\cite{bottou2012stochastic} in the self-supervised pretraining phase. We set the batch size to 128, the learning rates to 0.001,  and the temperature $\tau$ to 0.007. 

\textbf{Fine-tuning}: During the fine-tuning phase, we train DialogConv and other models using the Adam optimizer~\cite{kingma2014adam}. The learning rates are initialized to 1$e$-3, 5$e$-4, 1$e$-4, 5$e$-5 and 1$e$-5 via a multi-step strategy. The batch size is set to 32 for the MuTual dataset and 64 for the other datasets.
The values of the above hyperparameters are all fixed using the validation set.

\subsection{Results of Effectiveness}
Tables~\ref{compared_results} and~\ref{compared_results_2} report the test results of DialogConv and all compared models on the four datasets. While DialogConv does not achieve the best performance, the model attains near-optimal results in most cases. Furthermore, we calculate the confidence level ($p<0.05$) of DialogConv compared to BERT$_{base}$ (i.e., BERT-12-768), which shows that the results of DialogConv are credible.

As shown in Table~\ref{compared_results}, DialogConv outperforms most classic models such as DUA and DAM, and achieves comparable performance to MRFN on the Ubuntu dataset. DialogConv also outperforms other lightweight variants of BERT such as DBETR-6-768 (i.e., DistilBERT-6-768) and TBERT-6-768 (i.e., TinyBETR-6-768). When pretrained with contrastive learning, DialogConv performs close to BERT-12-768 and even outperforms BERT-12-768 on R10@2. On the Douban dataset, the performance of DialogConv is 2.3\% lower than the best result on R10@1. However, the performance of pretrained DialogConv can achieve near-optimal results. 

In Table~\ref{compared_results_2}, compared to BERT-12-768, DialogConv has a huge advantage of 21.7\% on R10@1 and 7.5\% higher on R10@2, is much better than other variants of BERT. We will discuss this phenomenon in Section~\ref{dis}. DialogConv outperforms some classic retrieval-based dialogue models\footnote{\url{https://nealcly.github.io/MuTual-leaderboard/}} such as DAM and MSN, and is close to some lightweight BERT variants such as DBERT-6-768 and BERT-4-512. Compared to BERT-12-768, DialogConv is 2.6\% lower on R10@1 on the MuTual. We believe that the lower performance of DialogConv on MuTual is caused by a limitation of DialogConv itself, which we will discuss in detail in Section~\ref{dis}.

\subsection{Model Efficiency}
To measure the complexity of our base model, we analyze the actual inference time of the model on CPU and GPU, as well as the number of parameters, as shown in Table~\ref{ppl}. DialogConv a huge speed advantage over other models, no matter on CPU or GPU.
For example, on the Ubuntu dataset, DialogConv $4.19\times$ to $115.67\times$ faster on CPU and $2.39\times$ to $11.64\times$ faster on GPU, and the average inference speed is $115.67\times$ faster on CPU and $11.64\times$ faster on GPU than other models.
On all four benchmark datasets, the inference speed of DialogConv is on average $79.39\times$ faster on CPU and $10.64\times$ faster on GPU compared to other models. Overall, the gain of inference speed ranges from $1.97\times$ to $40.61\times$ on GPU and from $3.47\times$ to $697.00\times$ on CPU. The CPU and GPU devices are described in the Implementation Details~\ref{ImpDe} subsection above.

The average number of parameters of DialogConv on the four benchmark datasets is 12.4 million, which is $2.8\times$ larger than BERT-2-128, $1.1\times$ than BERT-4-256, and comparable to TBERT-4-312. However, DialogConv has clear advantages in performance and inference time over these models. Compared to TBERT-6-768 and DBERT-7-768, the average number of parameters of DialogConv is $4.9\times$ and $5.1\times$ smaller, respectively. Compared with BERT-12-768, the average number of parameters of DialogConv on four datasets is $8.5\times$ smaller. As compared to the classic models DUA, DAM, IOI and MSN, DialogConv needs approximately $3.5\times$ less parameters. Overall, DialogConv achieves promising results in both performance and inference time, but relies on generally less parameters. 
The main reason is that convolutional structure enjoys the advantage of shared parameters, which make DialogConv have fewer parameters compared to other models based on RNN or attention mechanism.

\subsection{Ablation Study}
\begin{table}[t!]
\centering
\scriptsize
\begin{tabular}{r|p{0.5cm}p{0.5cm}p{0.5cm}|p{0.6cm}p{0.6cm}p{0.6cm}}
\toprule
   & \multicolumn{3}{c}{MuTual (English)} & \multicolumn{3}{|c}{ECD (Chinese)} \\ \midrule
   Setting & R@1 & R@2 & MRR & R10@1 & R10@2 & R10@5  \\ \midrule
   DialogConv & 0.614 & 0.825 & 0.778 & 0.833 & 0.901 & 0.988 \\
   -LocM  & 0.580 & 0.786 & 0.754 & 0.813 & 0.881 & 0.958\\
   -ConM & 0.577 & 0.801 & 0.759 & 0.806 & 0.823 & 0.919\\
   -DisM & 0.578 & 0.785 & 0.753  & 0.810 & 0.845 & 0.910\\
   -Agg & 0.573 & 0.783 & 0.750 & 0.804 & 0.824 & 0.870\\ \bottomrule
\end{tabular}
\caption{Ablation results on validation set.} 
\label{module_ablation_result}
\end{table}

Table~\ref{module_ablation_result} reports the result of module ablation. 1) \textbf{-LocM} removes the local matching layer; 2) \textbf{-ConM} removes the context matching layer; 3) \textbf{-DisM} removes the discourse matching layer; 4) \textbf{-Agg} replaces the aggregation layer with max-pooling. 

We can observe that each submodule plays a vital role in DialogConv. Specifically, the local matching layer captures the features of each utterance by mixing the features from the embedding and word views. The context matching layer updates the matching features based on the entire dialogue context and response. The discourse matching layer extracts the dependencies between different local contexts composed of adjacent utterances. Comparatively, it seems that the local matching layer has a little less impact on the model performance than the other layers. We conjecture that the layer can only extract local features to some extent since convolution is better at capturing local features.

\subsection{Result Analysis and Discussion}
\label{dis}
BERT-12-768 is a representative BERT base version among other BERT variants. Therefore, we use it as the basic comparison model.
In Table~\ref{compared_results_2}, DialogConv has an absolute advantage of 21.7\% on R10@1 and 7.5\% on R10@2 compared with BERT-12-768 on ECD. We believe that there are three main reasons for this phenomenon. First, DialogConv focuses on  \textit{matching}, which can extract matching features from stereoscopic views. We visualize the convolution results of each layer of dialogConv as a heatmap (Figure~\ref{heatmap-ecomm} in Appendix~\ref{heatmapV}). According to the heatmap, DialogConv can capture key matching features between dialogue context and response. The local matching layer mainly focuses on the features between words. This is because we use $1\times 1$ convolution in the conv@1 and conv@2 layers while matching features appear between several overlapping words in the two layers. When we use larger convolution kernels, DialogConv starts to focus on matching features between phrases. A similar phenomenon can be observed in the context matching layer. We can see that after the local and global features are extracted by the discourse matching layer, some important features are clearly captured. Second, for ECD, the average overlap of keywords between response and context reaches about 40\%, which is beneficial for DialogConv to extract matching features from multi-view stereos. Third, ECD is a dataset in the domain of e-commerce. The domain-specific performance of BERT-12-768 may be mediocre because the pre-training corpora of BERT-12-768 is domain-agnostic.

In Table~\ref{compared_results_2}, DialogConv achieves relatively insufficient performance on MuTual. We believe the main reason is that Mutual is a human-labeled reasoning dataset for multi-turn dialogues. However, DialogConv focuses on matching between dialogue context and response, and lacks reasoning ability. Therefore, DialogConv cannot make a correct predictions on reasoning-oriented examples in MuTual. Figure~\ref{heatmap-mutual} (in Appendix~\ref{heatmapV}) demonstrates the convolutional heatmap visualization of DialogConv on MuTual. According to the heatmap, DialogConv erroneously focuses on the features of \textit{"and"}, \textit{"their"} and \textit{"in"} in the dialogue context, and dose not consider \textit{"5"} and \textit{"7"} as key features.

DialogConv can effectively exploit the dependencies between different local contexts composed of adjacent utterances. To reveal its capabilities in this regard, we perturb the dialogue structure by randomly perturbing the dialogue context and report the results in Table~\ref{discourse_result_1} and Table~\ref{discourse_result_2} (in Appendix). We can see that the performance of DialogConv degrades to varying degrees on the four benchmark datasets. Specifically, the performance drops by 12.9\% R10@1 on Ubuntu, 12\% R10@1 on Douban, 17.7\% R10@1 on ECD, and 7.4\% R@1 on MuTual. We speculate that the dialogue structure contains the dependencies between different local contexts, which is important for multi-turn response selection. When perturbing the dialog strecture, the dependencies between local contexts will be severely broken, resulting in performance degradation of DialogConv.

\section{Conclusion}
In this paper, we propose DialogConv, a multi-view lightweight architecture based exclusively on CNN. DialogConv conducts convolutions on embedding, word, and utterance views to capture matching features. Experiment results show that DialogConv has fewer parameters, is faster, and requires less computing resources to achieve competitive results on four benchmark datasets. DialogConv provides a valuable reference for the dialogue system being deployed in real-world scenarios.
\section{Limitations}
Although our work can achieve competitive results with less computing resources, we acknowledge some limitations of our study. Firstly, DialogConv focuses on matching, resulting in insufficient reasoning ability. Therefore, DialogConv has a lot of room for improvement in the performance of dialogue reasoning (on the MuTual dataset). Secondly, we did not explore the performance of deep DialogConv. Our study mainly focuses on designing a lightweight model, ignoring the potential heavy-duty DialogConv under the blessing of large-scale training corpora. We will explore the performance potential of deep DialogConv in future work.
\section{Acknowledgement}
We would like to thank the reviewers for their constructive comments. The project is supported by the National Natural Science Foundation of China (61872074, 62172086, 62106039, 62272092).

\bibliography{anthology,custom}

\begin{thebibliography}{47}
\expandafter\ifx\csname natexlab\endcsname\relax\def\natexlab#1{#1}\fi

\bibitem[{{Baeza-Yates} and {Ribeiro-Neto}(1999)}]{Ricardo2016}
Ricardo~A. {Baeza-Yates} and Berthier {Ribeiro-Neto}. 1999.
\newblock \emph{Modern Information Retrieval}.

\bibitem[{Bahdanau et~al.(2015)Bahdanau, Cho, and Bengio}]{bahdanau2014neural}
Dzmitry Bahdanau, Kyung~Hyun Cho, and Yoshua Bengio. 2015.
\newblock Neural machine translation by jointly learning to align and
  translate.
\newblock In \emph{3rd International Conference on Learning Representations,
  ICLR 2015}.

\bibitem[{Bottou(2012)}]{bottou2012stochastic}
L{\'e}on Bottou. 2012.
\newblock Stochastic gradient descent tricks.
\newblock In \emph{Neural networks: Tricks of the trade}, pages 421--436.
  Springer.

\bibitem[{Chen et~al.(2017)Chen, Sun, Shi, and Hong}]{chen2017sampling}
Ting Chen, Yizhou Sun, Yue Shi, and Liangjie Hong. 2017.
\newblock On sampling strategies for neural network-based collaborative
  filtering.
\newblock In \emph{Proceedings of the 23rd ACM SIGKDD International Conference
  on Knowledge Discovery and Data Mining}, pages 767--776.

\bibitem[{Cho et~al.(2014)Cho, van Merrienboer, G{\"u}l{\c{c}}ehre, Bahdanau,
  Bougares, Schwenk, and Bengio}]{cho2014learning}
Kyunghyun Cho, Bart van Merrienboer, {\c{C}}aglar G{\"u}l{\c{c}}ehre, Dzmitry
  Bahdanau, Fethi Bougares, Holger Schwenk, and Yoshua Bengio. 2014.
\newblock Learning phrase representations using rnn encoder-decoder for
  statistical machine translation.
\newblock In \emph{EMNLP}.

\bibitem[{Cui et~al.(2020)Cui, Wu, Liu, Zhang, and Zhou}]{cui2020mutual}
Leyang Cui, Yu~Wu, Shujie Liu, Yue Zhang, and Ming Zhou. 2020.
\newblock Mutual: A dataset for multi-turn dialogue reasoning.
\newblock In \emph{Proceedings of the 58th Annual Meeting of the Association
  for Computational Linguistics}, pages 1406--1416.

\bibitem[{Dai et~al.(2021)Dai, Liu, Le, and Tan}]{dai2021coatnet}
Zihang Dai, Hanxiao Liu, Quoc Le, and Mingxing Tan. 2021.
\newblock Coatnet: Marrying convolution and attention for all data sizes.
\newblock \emph{Advances in Neural Information Processing Systems}, 34.

\bibitem[{{Devlin} et~al.(2019){Devlin}, {Chang}, {Lee}, and
  {Toutanova}}]{devlin2018bert}
Jacob {Devlin}, Ming-Wei {Chang}, Kenton {Lee}, and Kristina {Toutanova}. 2019.
\newblock Bert: Pre-training of deep bidirectional transformers for language
  understanding.
\newblock In \emph{NAACL-HLT 2019: Annual Conference of the North American
  Chapter of the Association for Computational Linguistics}, pages 4171--4186.

\bibitem[{Gao et~al.(2019)Gao, Cheng, Zhao, Zhang, Yang, and
  Torr}]{gao2019res2net}
Shanghua Gao, Ming-Ming Cheng, Kai Zhao, Xin-Yu Zhang, Ming-Hsuan Yang, and
  Philip~HS Torr. 2019.
\newblock Res2net: A new multi-scale backbone architecture.
\newblock \emph{IEEE transactions on pattern analysis and machine
  intelligence}.

\bibitem[{Gu et~al.(2020)Gu, Li, Liu, Ling, Su, Wei, and Zhu}]{gu2020speaker}
Jia-Chen Gu, Tianda Li, Quan Liu, Zhen-Hua Ling, Zhiming Su, Si~Wei, and
  Xiaodan Zhu. 2020.
\newblock Speaker-aware bert for multi-turn response selection in
  retrieval-based chatbots.
\newblock In \emph{Proceedings of the 29th ACM International Conference on
  Information \& Knowledge Management}, pages 2041--2044.

\bibitem[{Gu et~al.(2019)Gu, Ling, and Liu}]{gu2019interactive}
Jia-Chen Gu, Zhen-Hua Ling, and Quan Liu. 2019.
\newblock Interactive matching network for multi-turn response selection in
  retrieval-based chatbots.
\newblock In \emph{Proceedings of the 28th ACM International Conference on
  Information and Knowledge Management}, pages 2321--2324.

\bibitem[{Hadsell et~al.(2006)Hadsell, Chopra, and LeCun}]{1640964}
R.~Hadsell, S.~Chopra, and Y.~LeCun. 2006.
\newblock \href {https://doi.org/10.1109/CVPR.2006.100} {Dimensionality
  reduction by learning an invariant mapping}.
\newblock In \emph{2006 IEEE Computer Society Conference on Computer Vision and
  Pattern Recognition (CVPR'06)}, volume~2, pages 1735--1742.

\bibitem[{He et~al.(2016)He, Zhang, Ren, and Sun}]{he2016deep}
Kaiming He, Xiangyu Zhang, Shaoqing Ren, and Jian Sun. 2016.
\newblock Deep residual learning for image recognition.
\newblock In \emph{Proceedings of the IEEE conference on computer vision and
  pattern recognition}, pages 770--778.

\bibitem[{Hendrycks and Gimpel(2016)}]{hendrycks2016gaussian}
Dan Hendrycks and Kevin Gimpel. 2016.
\newblock Gaussian error linear units (gelus).
\newblock \emph{arXiv preprint arXiv:1606.08415}.

\bibitem[{Hochreiter and Schmidhuber(1997)}]{hochreiter1997long}
Sepp Hochreiter and J{\"u}rgen Schmidhuber. 1997.
\newblock Long short-term memory.
\newblock \emph{Neural computation}, 9(8):1735--1780.

\bibitem[{Jiao et~al.(2020)Jiao, Yin, Shang, Jiang, Chen, Li, Wang, and
  Liu}]{jiao2020tinybert}
Xiaoqi Jiao, Yichun Yin, Lifeng Shang, Xin Jiang, Xiao Chen, Linlin Li, Fang
  Wang, and Qun Liu. 2020.
\newblock Tinybert: Distilling bert for natural language understanding.
\newblock In \emph{Findings of the Association for Computational Linguistics:
  EMNLP 2020}, pages 4163--4174.

\bibitem[{Kingma and Ba(2015)}]{kingma2014adam}
Diederik~P. Kingma and Jimmy Ba. 2015.
\newblock Adam: A method for stochastic optimization.
\newblock In \emph{International Conference on Learning Representations
  (Poster)}.

\bibitem[{Krizhevsky et~al.(2012)Krizhevsky, Sutskever, and
  Hinton}]{krizhevsky2012imagenet}
Alex Krizhevsky, Ilya Sutskever, and Geoffrey~E Hinton. 2012.
\newblock Imagenet classification with deep convolutional neural networks.
\newblock \emph{Advances in neural information processing systems},
  25:1097--1105.

\bibitem[{{Lan} et~al.(2020){Lan}, {Chen}, {Goodman}, {Gimpel}, {Sharma}, and
  {Soricut}}]{Lan2020}
Zhenzhong {Lan}, Mingda {Chen}, Sebastian {Goodman}, Kevin {Gimpel}, Piyush
  {Sharma}, and Radu {Soricut}. 2020.
\newblock Albert: A lite bert for self-supervised learning of language
  representations.
\newblock In \emph{ICLR 2020 : Eighth International Conference on Learning
  Representations}.

\bibitem[{Li et~al.(2021)Li, Zhang, Fei, Feng, and Zhou}]{li2021conversations}
Zekang Li, Jinchao Zhang, Zhengcong Fei, Yang Feng, and Jie Zhou. 2021.
\newblock Conversations are not flat: Modeling the dynamic information flow
  across dialogue utterances.
\newblock In \emph{Proceedings of the 59th Annual Meeting of the Association
  for Computational Linguistics and the 11th International Joint Conference on
  Natural Language Processing (Volume 1: Long Papers)}, pages 128--138.

\bibitem[{Liu and Liu(2021)}]{liu2021simcls}
Yixin Liu and Pengfei Liu. 2021.
\newblock Simcls: A simple framework for contrastive learning of abstractive
  summarization.
\newblock In \emph{Proceedings of the 59th Annual Meeting of the Association
  for Computational Linguistics and the 11th International Joint Conference on
  Natural Language Processing (Volume 2: Short Papers)}, pages 1065--1072.

\bibitem[{Liu et~al.(2021)Liu, Feng, Wang, Song, Ren, and Zhang}]{liu2021graph}
Yongkang Liu, Shi Feng, Daling Wang, Kaisong Song, Feiliang Ren, and Yifei
  Zhang. 2021.
\newblock \href {https://ojs.aaai.org/index.php/AAAI/article/view/17585} {A
  graph reasoning network for multi-turn response selection via customized
  pre-training}.
\newblock \emph{Proceedings of the AAAI Conference on Artificial Intelligence},
  35(15):13433--13442.

\bibitem[{Lowe et~al.(2015)Lowe, Pow, Serban, and Pineau}]{lowe2015ubuntu}
Ryan Lowe, Nissan Pow, Iulian~Vlad Serban, and Joelle Pineau. 2015.
\newblock The ubuntu dialogue corpus: A large dataset for research in
  unstructured multi-turn dialogue systems.
\newblock In \emph{Proceedings of the 16th Annual Meeting of the Special
  Interest Group on Discourse and Dialogue}, pages 285--294.

\bibitem[{Lu et~al.(2019)Lu, Zhang, Xie, Ling, Zhou, and
  Xu}]{lu2019constructing}
Junyu Lu, Chenbin Zhang, Zeying Xie, Guang Ling, Tom~Chao Zhou, and Zenglin Xu.
  2019.
\newblock Constructing interpretive spatio-temporal features for multi-turn
  responses selection.
\newblock In \emph{Proceedings of the 57th Annual Meeting of the Association
  for Computational Linguistics}, pages 44--50.

\bibitem[{Sanh et~al.(2019)Sanh, Debut, Chaumond, and
  Wolf}]{sanh2019distilbert}
Victor Sanh, Lysandre Debut, Julien Chaumond, and Thomas Wolf. 2019.
\newblock Distilbert, a distilled version of bert: smaller, faster, cheaper and
  lighter.
\newblock \emph{arXiv preprint arXiv:1910.01108}.

\bibitem[{Sankar et~al.(2019)Sankar, Subramanian, Pal, Chandar, and
  Bengio}]{sankar2019neural}
Chinnadhurai Sankar, Sandeep Subramanian, Christopher Pal, Sarath Chandar, and
  Yoshua Bengio. 2019.
\newblock Do neural dialog systems use the conversation history effectively? an
  empirical study.
\newblock In \emph{Proceedings of the 57th Annual Meeting of the Association
  for Computational Linguistics}, pages 32--37.

\bibitem[{{Seo} et~al.(2017){Seo}, {Kembhavi}, {Farhadi}, and
  {Hajishirzi}}]{seo2017bidirectional}
Minjoon {Seo}, Aniruddha {Kembhavi}, Ali {Farhadi}, and Hannaneh {Hajishirzi}.
  2017.
\newblock Bidirectional attention flow for machine comprehension.
\newblock In \emph{ICLR 2017 : International Conference on Learning
  Representations 2017}.

\bibitem[{Szegedy et~al.(2015)Szegedy, Liu, Jia, Sermanet, Reed, Anguelov,
  Erhan, Vanhoucke, and Rabinovich}]{szegedy2015going}
Christian Szegedy, Wei Liu, Yangqing Jia, Pierre Sermanet, Scott Reed, Dragomir
  Anguelov, Dumitru Erhan, Vincent Vanhoucke, and Andrew Rabinovich. 2015.
\newblock Going deeper with convolutions.
\newblock In \emph{Proceedings of the IEEE conference on computer vision and
  pattern recognition}, pages 1--9.

\bibitem[{Tan et~al.(2015)Tan, Santos, Xiang, and Zhou}]{tan2015lstm}
Ming Tan, Cicero~dos Santos, Bing Xiang, and Bowen Zhou. 2015.
\newblock Lstm-based deep learning models for non-factoid answer selection.
\newblock \emph{arXiv preprint arXiv:1511.04108}.

\bibitem[{Tao et~al.(2021)Tao, Feng, Yan, Wu, and Jiang}]{tao2021survey}
Chongyang Tao, Jiazhan Feng, Rui Yan, Wei Wu, and Daxin Jiang. 2021.
\newblock A survey on response selection for retrieval-based dialogues.
\newblock IJCAI.

\bibitem[{{Tao} et~al.(2019){Tao}, wei {wu}, {Xu}, {Hu}, {Zhao}, and
  {Yan}}]{Tao2020}
Chongyang {Tao}, wei {wu}, Can {Xu}, Wenpeng {Hu}, Dongyan {Zhao}, and Rui
  {Yan}. 2019.
\newblock One time of interaction may not be enough: Go deep with an
  interaction-over-interaction network for response selection in dialogues.
\newblock In \emph{ACL 2019 : The 57th Annual Meeting of the Association for
  Computational Linguistics}, pages 1--11.

\bibitem[{Tao et~al.(2019)Tao, Wu, Xu, Hu, Zhao, and Yan}]{tao2019multi}
Chongyang Tao, Wei Wu, Can Xu, Wenpeng Hu, Dongyan Zhao, and Rui Yan. 2019.
\newblock Multi-representation fusion network for multi-turn response selection
  in retrieval-based chatbots.
\newblock In \emph{Proceedings of the twelfth ACM international conference on
  web search and data mining}, pages 267--275.

\bibitem[{Vaswani et~al.(2017)Vaswani, Shazeer, Parmar, Uszkoreit, Jones,
  Gomez, Kaiser, and Polosukhin}]{vaswani2017attention}
Ashish Vaswani, Noam Shazeer, Niki Parmar, Jakob Uszkoreit, Llion Jones,
  Aidan~N Gomez, {\L}ukasz Kaiser, and Illia Polosukhin. 2017.
\newblock Attention is all you need.
\newblock \emph{Advances in neural information processing systems}, 30.

\bibitem[{Voorhees et~al.(1999)}]{voorhees1999trec}
Ellen~M Voorhees et~al. 1999.
\newblock The trec-8 question answering track report.
\newblock In \emph{Trec}, volume~99, pages 77--82.

\bibitem[{Wan et~al.(2016)Wan, Lan, Guo, Xu, Pang, and Cheng}]{wan2016deep}
Shengxian Wan, Yanyan Lan, Jiafeng Guo, Jun Xu, Liang Pang, and Xueqi Cheng.
  2016.
\newblock A deep architecture for semantic matching with multiple positional
  sentence representations.
\newblock In \emph{Proceedings of the AAAI Conference on Artificial
  Intelligence}, 1.

\bibitem[{Wang and Jiang(2016)}]{wang2016machine}
Shuohang Wang and Jing Jiang. 2016.
\newblock Machine comprehension using match-lstm and answer pointer.
\newblock \emph{arXiv preprint arXiv:1608.07905}.

\bibitem[{Wu et~al.(2016)Wu, Wu, Xing, Zhou, and Li}]{wu2016sequential}
Yu~Wu, Wei Wu, Chen Xing, Ming Zhou, and Zhoujun Li. 2016.
\newblock Sequential matching network: A new architecture for multi-turn
  response selection in retrieval-based chatbots.
\newblock \emph{arXiv preprint arXiv:1612.01627}.

\bibitem[{Wu et~al.(2017)Wu, Wu, Xing, Zhou, and Li}]{wu2017sequential}
Yu~Wu, Wei Wu, Chen Xing, Ming Zhou, and Zhoujun Li. 2017.
\newblock Sequential matching network: A new architecture for multi-turn
  response selection in retrieval-based chatbots.
\newblock In \emph{Proceedings of the 55th Annual Meeting of the Association
  for Computational Linguistics (Volume 1: Long Papers)}, pages 496--505.

\bibitem[{Wu et~al.(2019)Wu, Liu, Lin, Lin, and Han}]{wu2019lite}
Zhanghao Wu, Zhijian Liu, Ji~Lin, Yujun Lin, and Song Han. 2019.
\newblock Lite transformer with long-short range attention.
\newblock In \emph{International Conference on Learning Representations}.

\bibitem[{Yan et~al.(2016)Yan, Song, and Wu}]{yan2016learning}
Rui Yan, Yiping Song, and Hua Wu. 2016.
\newblock Learning to respond with deep neural networks for retrieval-based
  human-computer conversation system.
\newblock In \emph{Proceedings of the 39th International ACM SIGIR conference
  on Research and Development in Information Retrieval}, pages 55--64.

\bibitem[{{Yu} et~al.(2018){Yu}, {Dohan}, {Luong}, {Zhao}, {Chen}, {Norouzi},
  and {Le}}]{yu2018qanet}
Adams~Wei {Yu}, David {Dohan}, Minh-Thang {Luong}, Rui {Zhao}, Kai {Chen},
  Mohammad {Norouzi}, and Quoc~V. {Le}. 2018.
\newblock Qanet: Combining local convolution with global self-attention for
  reading comprehension.
\newblock In \emph{International Conference on Learning Representations}.

\bibitem[{Yuan et~al.(2019)Yuan, Zhou, Li, Lv, Zhu, Han, and
  Hu}]{yuan2019multi}
Chunyuan Yuan, Wei Zhou, Mingming Li, Shangwen Lv, Fuqing Zhu, Jizhong Han, and
  Songlin Hu. 2019.
\newblock Multi-hop selector network for multi-turn response selection in
  retrieval-based chatbots.
\newblock In \emph{Proceedings of the 2019 Conference on Empirical Methods in
  Natural Language Processing and the 9th International Joint Conference on
  Natural Language Processing (EMNLP-IJCNLP)}, pages 111--120.

\bibitem[{Zhang et~al.(2018)Zhang, Li, Zhu, Zhao, and Liu}]{zhang2018modeling}
Zhuosheng Zhang, Jiangtong Li, Pengfei Zhu, Hai Zhao, and Gongshen Liu. 2018.
\newblock Modeling multi-turn conversation with deep utterance aggregation.
\newblock In \emph{Proceedings of the 27th International Conference on
  Computational Linguistics}, pages 3740--3752.

\bibitem[{Zhang and Zhao(2021)}]{zhang2021advances}
Zhuosheng Zhang and Hai Zhao. 2021.
\newblock Advances in multi-turn dialogue comprehension: A survey.
\newblock \emph{arXiv preprint arXiv:2103.03125}.

\bibitem[{Zhou et~al.(2016)Zhou, Dong, Wu, Zhao, Yu, Tian, Liu, and
  Yan}]{zhou2016multi}
Xiangyang Zhou, Daxiang Dong, Hua Wu, Shiqi Zhao, Dianhai Yu, Hao Tian, Xuan
  Liu, and Rui Yan. 2016.
\newblock Multi-view response selection for human-computer conversation.
\newblock In \emph{Proceedings of the 2016 Conference on Empirical Methods in
  Natural Language Processing}, pages 372--381.

\bibitem[{{Zhou} et~al.(2018){Zhou}, {Li}, {Dong}, {Liu}, {Chen}, {Zhao}, {Yu},
  and {Wu}}]{Zhou2018}
Xiangyang {Zhou}, Lu~{Li}, Daxiang {Dong}, Yi~{Liu}, Ying {Chen}, Wayne~Xin
  {Zhao}, Dianhai {Yu}, and Hua {Wu}. 2018.
\newblock Multi-turn response selection for chatbots with deep attention
  matching network.
\newblock In \emph{ACL 2018: 56th Annual Meeting of the Association for
  Computational Linguistics}, volume~1, pages 1118--1127.

\bibitem[{Zhou et~al.(2018)Zhou, Li, Dong, Liu, Chen, Zhao, Yu, and
  Wu}]{zhou2018multi}
Xiangyang Zhou, Lu~Li, Daxiang Dong, Yi~Liu, Ying Chen, Wayne~Xin Zhao, Dianhai
  Yu, and Hua Wu. 2018.
\newblock Multi-turn response selection for chatbots with deep attention
  matching network.
\newblock In \emph{Proceedings of the 56th Annual Meeting of the Association
  for Computational Linguistics (Volume 1: Long Papers)}, pages 1118--1127.

\end{thebibliography}
\bibliographystyle{acl_natbib}

\appendix
\section{Appendix}
\subsection{Baselines}
\label{baseline}
\textbf{MV-LSTM}~\cite{wan2016deep} is a semantic matching method based on LSTM. \textbf{QANET}~\cite{yu2018qanet} is a machine reading comprehension method based on CNN. \textbf{MH-LSTM}~\cite{wang2016machine} is an extractive machine reading comprehension model based on LSTM. \textbf{BIDAF}~\cite{seo2017bidirectional} is a machine reading comprehension model based on bi-directional attention flow. \textbf{Multi-View}~\cite{zhou2016multi} is a multi-turn dialogue retrieval-based method based on token view and utterance view. \textbf{DL2R} ~\cite{yan2016learning} is a multi-turn retrieval-based dialogue model based on sentence pair matching. \textbf{DUA}~\citep{zhang2018modeling} is a hierarchical interaction model based on attention mechanism. \textbf{DAM}~\citep{zhou2018multi} is a deep interaction method based on attention. \textbf{IMN}~\cite{gu2019interactive} is a retrieval-based dialogue model with bi-directional matching. \textbf{MRFN}~\cite{tao2019multi} is a retrieval-based dialogue model with multiple types of representations. \textbf{IoI}~\citep{Tao2020} is a retrieval-based dialogue model based on multiple interactions. \textbf{MSN}~\cite{yuan2019multi} is a retrieval-based dialogue model with multi-hop selector mechanism.

\textbf{BERT}~\citep{devlin2018bert} is an autoencoding language model based on transformer. Here we employ multiple BERT versions including BERT-2-128 (two layers with hidden size 128), BERT-4-256 (four layers with hidden size 256), BERT-4-512, BERT-8-512 and BERT-12-768. \textbf{TinyBERT}~\cite{jiao2020tinybert} is compressed BERT through a two-stage distillation technique. TinyBERT includes the officially releases TinyBERT-3-312 (short for TBERT-3-312) and TinyBERT-6-768 (short for TBERT-6-768) in Chinese and English. \textbf{DistilBERT}~\cite{sanh2019distilbert} is a distilled version of BERT. DistilBERT includes the officially releases DistilBERT-6-768 (short for DBERT-6-768) in Chinese and English. Notationwise, we use $X$ to represent the number of layers and $Y$ represent the hidden size in BERT-$X$-$Y$, DBERT-$X$-$Y$ and TBERT-$X$-$Y$.

\subsection{Heatmap Visualization}
\label{heatmapV}

\begin{table*}[t!]
\centering
\begin{tabular}{cll}
\hline
  & Chinese & English \\ \hline
  \multirow{3}{*}{context} 
  &\begin{CJK}{UTF8}{gbsn} \small{我 提前 问 你 了 吧} \end{CJK}& I asked you in advance.\\
  &\begin{CJK}{UTF8}{gbsn} \small{这 不是 质量 问题 哦} \end{CJK}& It's not a quality problem. \\
  &\begin{CJK}{UTF8}{gbsn} \small{我 要 小 的 又 不能 用 而且 这么 多包} \end{CJK}& I want small and unusable, and so many bags. \\ \hline
 \multirow{2}{*}{response} 
 &\begin{CJK}{UTF8}{gbsn} \small{非 质量 问题 退回 的} \end{CJK} & For non-quality problems returned \\
 &\begin{CJK}{UTF8}{gbsn} \small{运费 亲自 理} \end{CJK} & the freight will be handled by yourself  \\\hline
\end{tabular}
\caption{An example of a Chinese-English aligned of ECD dataset.} 
\label{zh-en}
\end{table*}

\begin{figure*}[t!]
\centering 
\includegraphics[width=0.91\textwidth]{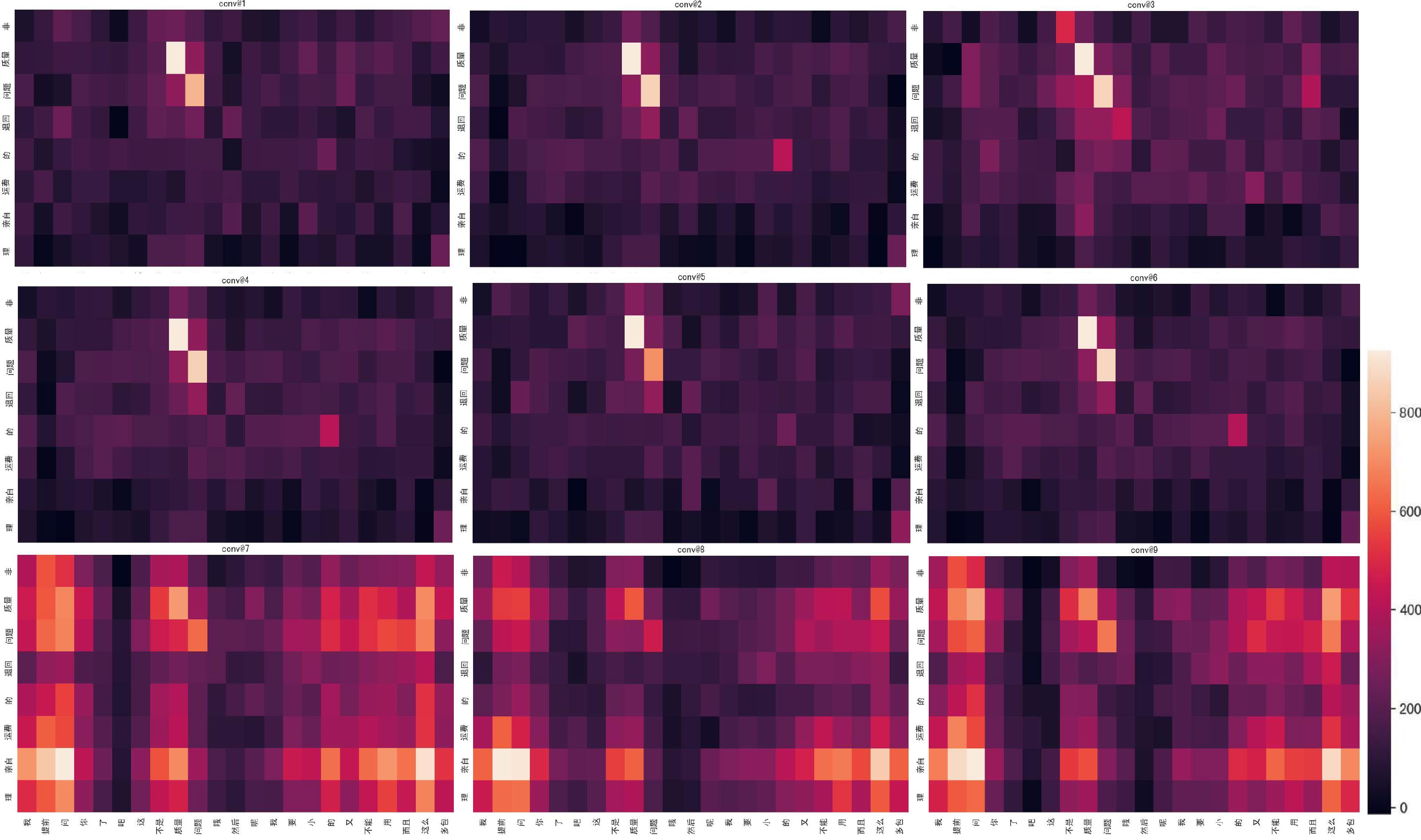}
\caption{An example of visualization heatmap from ECD. The conv@\textit{i} represents the i-th convolution operation in Figure~\ref{framework}.
The horizontal axis represents the dialogue history, and the vertical axis represents the response. The English translation refers to Table~\ref{zh-en}.}
\label{heatmap-ecomm}
\end{figure*}
\begin{figure*}[t!]
\centering 
\includegraphics[width=0.91\textwidth]{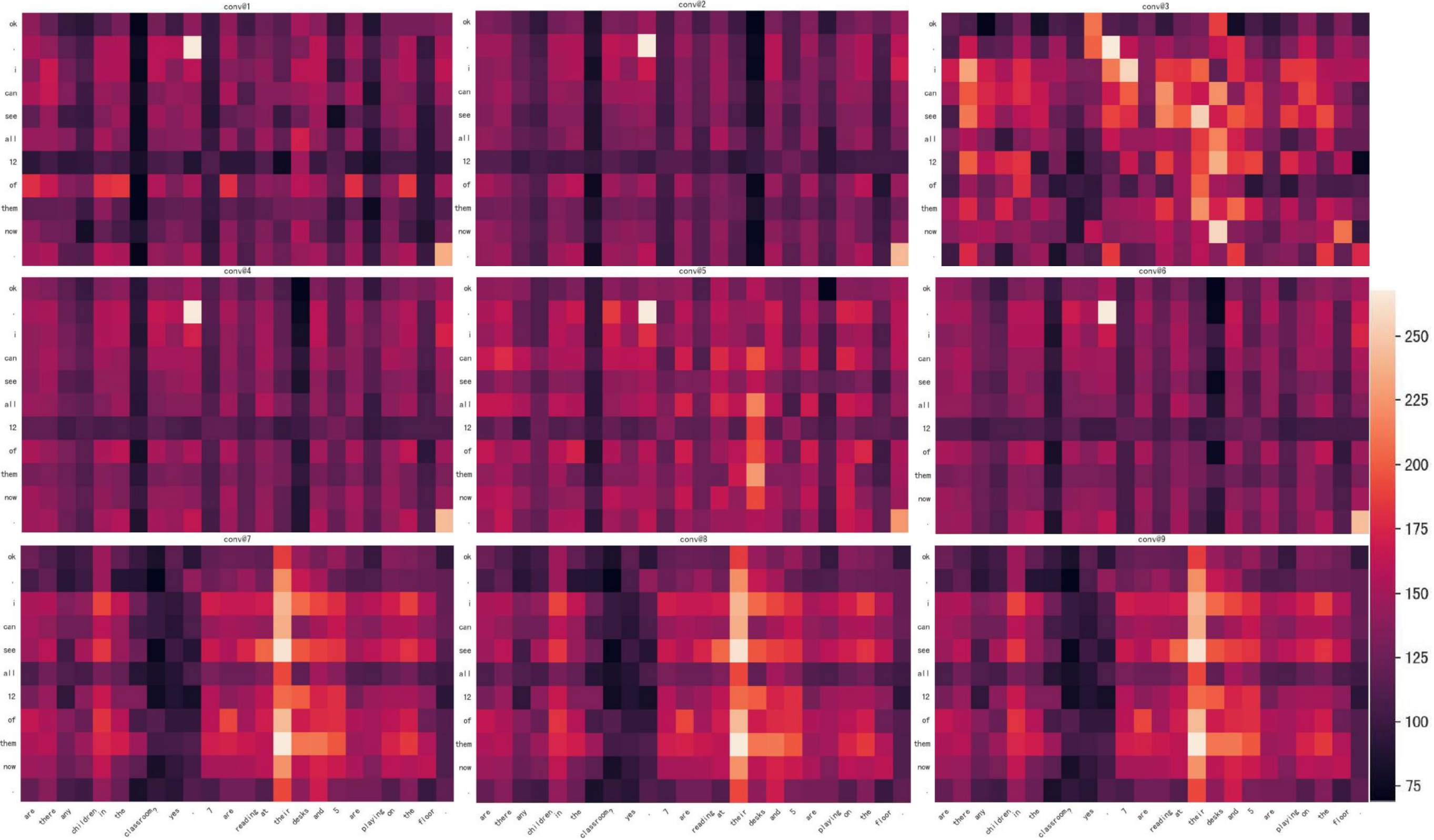}
\caption{An example of visualization heatmap from Mutual. The conv@\textit{i} represents the i-th convolution operation in Figure~\ref{framework}. The horizontal axis represents the dialogue history, and the vertical axis represents the response.}
\label{heatmap-mutual}
\end{figure*}

Figure~\ref{heatmap-ecomm} and Figure~\ref{heatmap-mutual} demonstrate example convolutional heatmap visualizations for each layer of DialogConv from datasets Mutual and ECD, respectively. Table~\ref{zh-en} demonstrates the comparison between Chinese and English of an example of Ecomm. We obtain the heatmaps in Figure~\ref{heatmap-mutual} and ~\ref{heatmap-ecomm} through visualizing the similarity matrix between response and dialogue context. The larger the value of the similarity matrix, the brighter the corresponding visualization result, and the more important the corresponding word is. According to Figure~\ref{heatmap-ecomm}, DialogConv can capture the key features in dialogue context and response such as \textit{"not"} (\begin{CJK}{UTF8}{gbsn}不是\end{CJK} or \begin{CJK}{UTF8}{gbsn}非\end{CJK} ), \textit{"quality"} (\begin{CJK}{UTF8}{gbsn}质量\end{CJK}), \textit{"problem"} (\begin{CJK}{UTF8}{gbsn}问题\end{CJK}). We can conclude that DialogConv makes decision based on the matching features between dialogue context and responses. According to Figure~\ref{heatmap-mutual}, DialogConv mistakenly considers \textit{"their"}, \textit{"them"}, \textit{"see"}, and \textit{"i"} as important features and ignores the key features \textit{"5"} and \textit{"7"} in the dialogue context.

\begin{table*}[t!]
\centering
\begin{tabular}{l|ccc|cccccc}
\toprule
    & \multicolumn{3}{c}{Ubuntu (English)} & \multicolumn{6}{|c}{Douban (Chinese)} \\ \midrule
   Types & R10@1 & R10@2 & R10@5 & MAP &MRR & P@1 & R10@1 &R10@2 & R10@5  \\ \midrule
  Norm & 0.788 & 0.883 & 0.979 & 0.571 & 0.624 & 0.432 & 0.272 & 0.453 & 0.785 \\
  Rand & 0.659 & 0.781 & 0.948 & 0.422 & 0.458 & 0.267 & 0.152 & 0.270 & 0.630\\
  \bottomrule
\end{tabular}
\caption{Performance of DialogConv with normal/perturbed dialogue structure on Ubuntu and Douban. Norm represents the normal dialogue structure. Rand represents the perturbing dialogue structure by shuffling the dialogue context randomly.}
\label{discourse_result_1}
\end{table*}

\begin{table*}[t!]
\centering
\begin{tabular}{l|ccc|ccc}
\toprule
   & \multicolumn{3}{c}{ECD (Chinese)} & \multicolumn{3}{|c}{MuTual (English)} \\ \midrule
  Types & R10@1 & R10@2 & R10@5 & R@1 & R@2 & MRR   \\ \midrule
  Norm  & 0.827 & 0.889 & 0.962 & 0.602 & 0.834 & 0.769  \\
  Rand  & 0.650 & 0.780 & 0.946 & 0.528 & 0.761 & 0.686  \\
  \bottomrule
\end{tabular}
\caption{Performance of DialogConv with normal/perturbed dialogue structure on ECD and MuTual. Norm represents the normal dialogue structure. Rand represents the perturbing dialogue structure by shuffling the dialogue context randomly.} 
\label{discourse_result_2}
\end{table*}

\end{document}